\documentclass[lettersize,journal, twoside]{IEEEtran}
\usepackage{hyperref}

\usepackage{amsmath,amsfonts}
\usepackage{mathrsfs}
\usepackage{algorithmic}
\usepackage{algorithm}
\usepackage{array}
\usepackage[caption=false,font=normalsize,labelfont=sf,textfont=sf]{subfig}
\usepackage{textcomp}
\usepackage{stfloats}
\usepackage{url}
\usepackage{verbatim}
\usepackage[capitalise,nameinlink]{cleveref}
\crefname{figure}{Fig.}{Figs.}
\Crefname{figure}{Fig.}{Figs.}
\crefname{table}{Tab.}{Tabs.}
\Crefname{table}{Tab.}{Tabs.}
\crefname{equation}{Eq.}{Eqs.}
\Crefname{equation}{Eq.}{Eqs.}

\usepackage{graphicx}
\usepackage{booktabs} 
\usepackage{cite}
\usepackage{makecell}
\usepackage{rotating}
\usepackage[table,xcdraw]{xcolor}
\usepackage{multirow, multicol}
\usepackage{booktabs} 
\usepackage{amsbsy} 
\usepackage{bm} 
\usepackage{arydshln} % 支持虚线的表格宏包
\usepackage{makecell} 
\usepackage{pifont}
\usepackage{xcolor}
\usepackage{enumitem}
\usepackage{caption}
\usepackage{multirow}
\usepackage{xcolor}
\usepackage{arydshln}

\begin{document}

\title{\fontsize{20}{5}\selectfont EGS-SLAM: RGB-D Gaussian Splatting SLAM with Events}

\author{
Siyu Chen$^{\ast}$,
Shenghai Yuan$^{\ast}$,
Thien-Minh Nguyen, \\
Zhuyu Huang,
Chenyang Shi,
Jin Jing,
Lihua Xie \textit{Fellow, IEEE}
\thanks{$^{\ast}$Equal contribution.}
\vspace{-25pt}
\thanks{$^{\dagger}$ This work is supported by any National Research Foundation, Singapore
under its Medium Sized Center for Advanced Robotics Technology Innovation.}
\thanks{Manuscript received: April 29, 2025; Revised: July 11, 2025; Accepted: August 4, 2025.  This paper was recommended for publication by Editor Javier Civera upon valuation of the Associate Editor and Reviewers' comments}
\thanks{Siyu Chen, Shenghai Yuan, Thien-Minh Nguyen, Lihua Xie are with the School of Electrical and Electronic Engineering, Nanyang Technological University, Singapore. {\tt\small \{siyu010, shyuan, thienminh.nguyen, elhxie\}@ntu.edu.sg}}
\thanks{Zhuyu Huang, Chenyang Shi, Jin Jing is with the School of Instrumentation and Optoelectronics Engineering, Beihang University, Beijing, China.  {\tt\small \{shicy\}@buaa.edu.cn}}
\thanks{Digital Object Identifier (DOI):  see top of this page.}
}

% \IEEEpubid{0000--0000/00\$00.00~\copyright~2021 IEEE}
% Remember, if you use this you must call \IEEEpubidadjcol in the second
% column for its text to clear the IEEEpubid mark.

\maketitle

\begin{abstract}
Gaussian Splatting SLAM (GS-SLAM) offers a notable improvement over traditional SLAM methods, in enabling photorealistic 3D reconstruction that conventional approaches often struggle to achieve. However, existing GS-SLAM systems perform poorly under persistent and severe motion blur commonly encountered in real-world scenarios, leading to significantly degraded tracking accuracy and compromised 3D reconstruction quality. To address this limitation, we propose EGS-SLAM, a novel GS-SLAM framework that fuses event data with RGB-D inputs to simultaneously reduce motion blur in images and compensate for the sparse, discrete nature of event streams, enabling robust tracking and high-fidelity 3DGS reconstruction. Specifically, our system explicitly models the camera's continuous trajectory during exposure, supporting event and blur-aware tracking and mapping on a unified 3DGS scene. Furthermore, we introduce a learnable camera response function to align the dynamic ranges of events and images, along with a no-event loss to suppress ringing artifacts during reconstruction. We validate our approach on a new dataset comprising synthetic and real-world sequences with significant motion blur. Extensive experimental results demonstrate that EGS-SLAM consistently outperforms existing GS-SLAM systems in both trajectory accuracy and photorealistic 3DGS reconstruction.
The source code will be available at \url{https://github.com/Chensiyu00/EGS-SLAM}.
\end{abstract}


\section{Introduction}
Simultaneous Localization and Mapping (SLAM) is fundamental to robotic autonomy, enabling agents to estimate their pose and build environmental maps in unknown settings~\cite{liu2023relative, yuan2024large, li2025ua, pypose,UnmannedSystems2025}. While most visual SLAM systems~\cite{orbslam2, orbslam3, droidslam, sdpvo} achieve high localization accuracy, they struggle to reconstruct detailed geometry and photorealistic appearance. Recently, 3D Gaussian Splatting (3DGS) has been introduced into SLAM~\cite{monogs, gsslam, splatam, photoslam}, offering explicit scene representations with real-time rendering and richer reconstructions. However, these methods typically assume high-quality, blur-free images as input.

Nevertheless, under fast or continuous motion, conventional cameras often produce motion-blurred frames, leading to the loss of critical visual details. This degradation adversely affects both camera tracking and scene reconstruction, limiting the performance of 3DGS-based SLAM systems. Although existing SLAM approaches~\cite{i2slam, mbaslam} attempt to address this issue through image-only deblurring techniques, they remain ineffective under sustained or strong blur, as they fail to recover sufficient detail for reliable operation. Consequently, current high-accuracy, photorealistic SLAM systems struggle to operate reliably in environments affected by motion blur.

To address these limitations, incorporating additional sensing modalities is a promising direction. Event cameras offer a compelling alternative, as their microsecond-level asynchronous brightness sensing inherently avoids motion blur. In existing works~\cite{eventlineslam, ultimateslam}, event data has been employed in SLAM systems for robust camera tracking and in 3D Gaussian reconstruction~\cite{yu2024evagaussians, xiong2024event3dgs, han2024event} to recover sharp and detailed scenes under severe motion blur. To date, no existing work has integrated event information into the GS-SLAM framework to simultaneously enable online tracking under blurred conditions and the construction of high-fidelity 3D Gaussian maps.

Integrating event data into a GS-SLAM framework faces two main challenges. First, event streams are sparse and respond to per-pixel brightness changes, while image frames are captured at discrete intervals based on exposure. This inconsistency makes it challenging to fuse events and images for joint tracking and mapping within the 3DGS. Second, events and images differ significantly in their dynamic range: event data has inherently high dynamic range (HDR), whereas standard images have low dynamic range (LDR) and are constrained by exposure settings. This mismatch complicates fusion and the construction of a unified 3D Gaussian Splatting representation that effectively leverages both modalities.

To overcome these challenges, we propose an Event RGB-D GS-SLAM framework that uses event, image and depth within the camera’s exposure time for accurate tracking and mapping. We model the continuous camera trajectory during each exposure interval and use it to render blur-aware images and corresponding event maps from 3DGS. These rendered signals are compared against the accumulated event maps, obtained by integrating the raw event stream within the same interval, and the image for temporally consistent tracking and mapping. To bridge the dynamic range gap between HDR events and LDR images, we introduce a learnable Camera Response Function (CRF) that transforms both modalities into a shared intensity space. This unified design enables our system to robustly operate under motion blur, leveraging the complementary advantages of both sensing modalities.

Our contributions can be summarized as:
\begin{enumerate}[label=\textbullet] 
    \item To our knowledge, we present the first E-RGB-D GS-SLAM framework that incorporates event data alongside RGB and depth information. By jointly modeling these modalities within the camera's exposure time, our method enables robust tracking and mapping under severe blur.
    \item We design an event-aided tracker and mapper for GS-SLAM that operate on blurry images, events, and depth to achieve accurate tracking and high-quality mapping. In the tracker and mapper, we incorporate a learnable CRF to align HDR events with LDR images and introduce a no-event loss to suppress ringing artifacts.
    \item We construct a dataset containing both synthetic and real-world sequences with challenging motion blur. Extensive experiments show that our method outperforms existing GS-SLAM and classical baselines, both in camera localization and in reconstructing high-fidelity 3DGS.
\end{enumerate}

\section{Related Works}
This section presents deblur GS/NeRF, event-based GS/NeRF, and GS-SLAM which are relevant to this work.

\textbf{Deblur GS/NeRF}
A line of research tackles 3D GS/NeRF reconstruction from motion-blurred images.
Deblur-NeRF~\cite{deblurnerf} introduces a deformable sparse kernel to recover sharp NeRFs from blurry inputs.
BAD-NeRF~\cite{badnerf} models dynamic blur trajectories to reconstruct clean scenes, and BAD-GS~\cite{badgs} extends this idea to 3DGS, yielding faster training and rendering while preserving detail.
Seiskari \emph{et al.}~\cite{gsonmove} incorporate velocity from visual-inertial odometry to mitigate both motion-blur and rolling-shutter artifacts, whereas BARD-GS~\cite{lu2025bardgsblurawarereconstructiondynamic} jointly models camera and object motion for dynamic scenes.
Despite this progress, purely image-based methods still falter under extreme or persistent blur and all of these methods are offline.

\begin{figure*}
    \centering
    \includegraphics[width=0.9\linewidth]{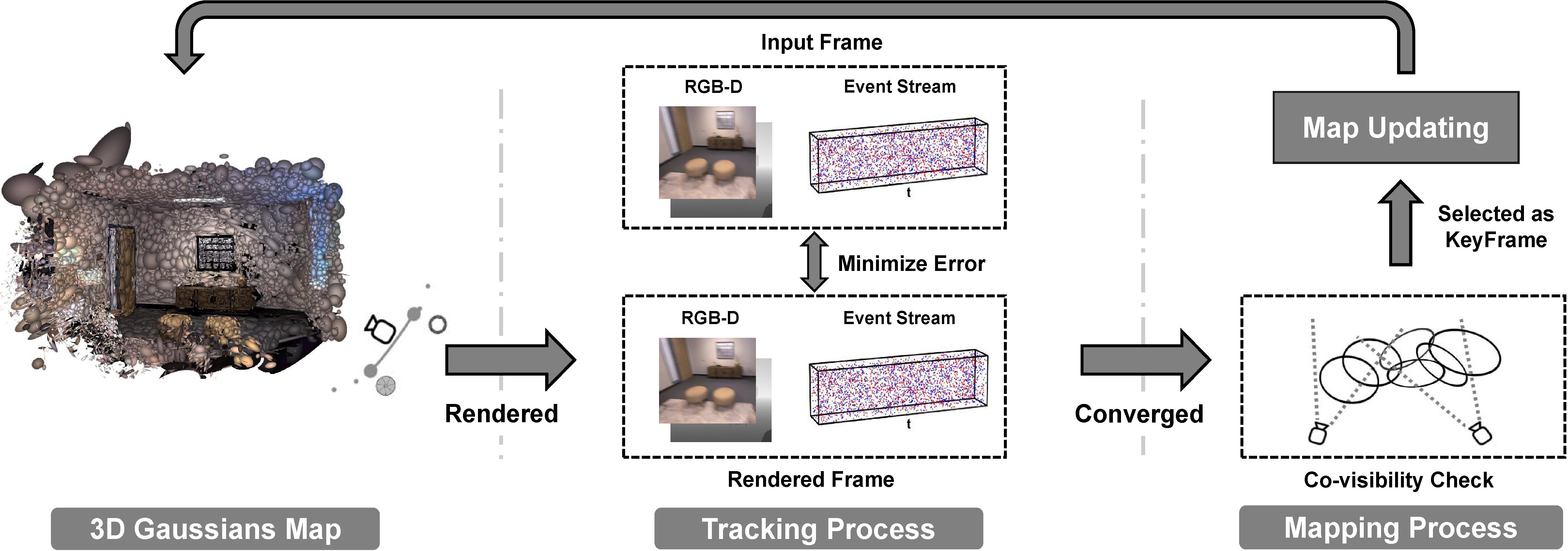}
    \caption{\textbf{The pipeline of Gaussian Splatting SLAM with Event.} Our system integrates event–image-depth tracking and mapping within a unified 3DGS map. For each frame, the pose is estimated by jointly rendering events, image and depth. Once converged, the mapping module updates the keyframe window and the 3DGS map if selected.}
    \label{fig1}
    \vspace{-0.8em}
\end{figure*}

\textbf{Event-based GS/NeRF.}
Event cameras provide asynchronous, blur-free measurements that are well suited for 3D reconstruction.
E\textsuperscript{2}NeRF~\cite{qi2023e2nerf} couples blur- and event-rendering losses to obtain sharp NeRFs from heavily blurred images.
Ev-DeblurNeRF~\cite{Ev-DeblurNeRF} learns an event-to-pixel response to denoise events and boost reconstruction quality, while E-NeRF~\cite{enerf} achieves event-only NeRF from a single sensor.
Event3DGS~\cite{xiong2024event3dgs} combines events with blur modeling for crisp 3DGS results, and IncEventGS~\cite{huang2024inceventgs} attains high-quality, pose-free 3DGS using events alone.
However, all of these methods (except IncEventGS) require offline initialization of camera poses via structure-from-motion (SfM) before optimizing the neural or Gaussian scene representation and therefore cannot support online reconstruction and localization.

\textbf{GS-SLAM}
Integrating 3DGS into SLAM has produced more photorealistic and efficient maps. GS-SLAM~\cite{gsslam}, SplaTAM~\cite{splatam}, and MonoGS~\cite{monogs} unify tracking and mapping within a single 3DGS representation, while PhotoSLAM~\cite{photoslam} reconstructs 3DGS from ORB-SLAM poses. These systems perform well on sharp inputs but severely degrade under motion blur. MBA-SLAM~\cite{mbaslam} embeds the camera-imaging process to handle blurred frames, and I\textsuperscript{2}-SLAM~\cite{i2slam} incorporates the camera-response function for improved mapping. Nonetheless, all of these are image-only methods, which remain vulnerable to continuous and intense motion blur due to their reliance on blurred frame observations.

To our knowledge, only one NeRF-based SLAM~\cite{eventnerfslam} considers combining the event and frames to achieve online tracking and mapping. Crucially, no prior work has yet integrated events and RGB images into GS-SLAM that  delivers online accurate tracking and high-quality 3DGS mapping.

\section{Method}
Our SLAM system is built around three tightly integrated components: (i) a unified 3DGS map, (ii) an event-image tracking process, and (iii) an incremental event-image mapping process, as illustrated in~\Cref{fig1}. The 3DGS map serves as the only scene representation for tracking and mapping. For each incoming frame, the tracking module jointly leverages the event stream, depth and image for tracking, estimating a continuous camera trajectory by rendering both signals from the 3DGS map over the frame's exposure interval. Once the pose of the incoming frame has converged, the mapping module evaluates its overlap with the latest keyframes to decide whether it should be selected as a new keyframe. If selected, it updates and maintains the current keyframe window and then updates the 3DGS map accordingly.

\subsection{3D Gaussian Splatting Map Representation} 
Following~\cite{gs}, a 3D scene is represented as a set of 3D Gaussians $\mathcal{G}$, each characterized by
a mean $\bm{\mu}_i\in \mathbb{R}^3$, a covariance matrix $\Sigma_i \in \mathbb{R}^{3\times3} = R_i s_i s_i^T R_i^T$ represented by a rotation matrix $R \in \mathrm{SO}(3)$ given by a unit quaternion $q_i$ and a scale $s_i \in \mathbb{R}^3$ , an opacity $o_i \in \mathbb{R}$, and a color $\bm{c}_i \in \mathbb{R}^{3}$. For efficiency, the spherical harmonic representation is omitted, as done in~\cite{monogs}. To render an image from these 3D Gaussians, we first project them onto the image plane using the camera pose:
\begin{equation}
    \label{eq1}
    \begin{aligned}
        \hat{\bm{\mu}}_i = \pi(T_{cw}, \bm{\mu}_i); \hat{\Sigma}_i = J R_{cw} \Sigma_i R_{cw}^T J^T,
    \end{aligned}
\end{equation}
where $\hat{\mu}_i$ is the projected mean, $\pi(\cdot)$ represents the projection function, and $T_{cw}$ is the world-to-camera transformation.
$\hat{\Sigma}_i$ is the projected covariance. $J$ is the Jacobian of the affine transformation, 
and $R_{cw}$ is the rotation matrix of $T_{cw}$. The rendered color $\bm{\mathcal{I}}(\bm{u})$ and depth $\bm{\mathcal{D}}(\bm{u})$ of the pixel at the $\bm{u}$ position are then computed by the $\bm{\alpha}$-blending of all reprojected Gaussians overlapping on the pixel, sorted by the depth:
\begin{equation}\label{eq2}
\footnotesize
    \begin{aligned}
       \bm{\mathcal{I}}(\bm{u}) = \sum_{i \in \mathcal{N}} \bm{c}_{i} \alpha_i \prod_{j=1}^{i-1} (1-\alpha_j);
       \bm{\mathcal{D}}(\bm{u}) = \sum_{i \in \mathcal{N}} d_{i} \alpha_i \prod_{j=1}^{i-1} (1-\alpha_j),
    \end{aligned}
\end{equation}
where $d_i$ denotes the projected depth of the center of the $i$-th 3D Gaussian 
and $\alpha_i=o_i e^{-\frac{1}{2}(\bm{u} -\bm{\hat{\mu}}_i)^T \hat{\Sigma}^{-1}_i (\bm{u} -\bm{\hat{\mu}}_i)}$. 
For simplicity, we denote $T_{cw}$ as $\bm{T}$ in the rest of the paper.

\subsection{Camera Motion Modeling during Exposure} 
To model the camera's continuous motion trajectory within a single exposure duration, we assume the motion is linear and express the pose at any instant $\eta \in [0, \tau]$ by interpolating between the start pose $\mathbf{T}_0$ and end pose $\mathbf{T}_\tau$:
\begin{equation}
    \label{eq4}
    \bm{T}(\eta) = \begin{bmatrix}
        \text{Slerp}(R_0, R_\tau, \frac{\eta}{\tau}) & (1-\frac{\eta}{\tau})\mathbf{t}_0 + \frac{\eta}{\tau}\mathbf{t}_\tau \\
        \mathbf{0} & 1
    \end{bmatrix},
\end{equation}
where $R_0, R_\tau \in \mathrm{SO}(3)$ and $\mathbf{t}_0, \mathbf{t}_\tau \in \mathbb{R}^3$ are the rotation and translation components of $\mathbf{T}_0$ and $\mathbf{T}_\tau$, representing the camera poses at the start ($\eta = 0$) and end ($\eta = \tau$) of exposure. $\text{Slerp}(\cdot)$ denotes spherical linear interpolation in $\mathrm{SO}(3)$.

\subsection{Blur-Aware Tracking with Event}
\begin{figure}[t]
    \centering
    \includegraphics[width=0.95\linewidth]{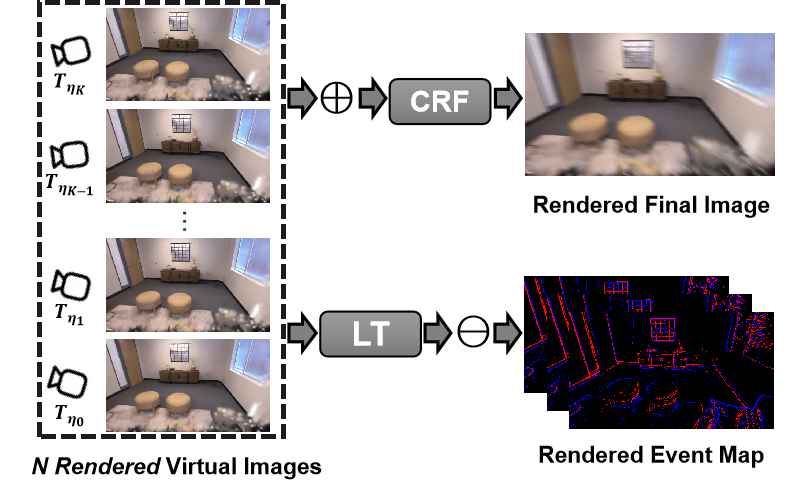}
    \caption{\textbf{Illustration of blur-aware image and event map rendering.} Rendered images along the trajectory are aggregated via a CRF for the final image. Event maps are generated by computing logarithmic brightness differences between consecutive rendered frames.}
    \label{fig:render}
\end{figure}

The tracking module in our SLAM system estimates the optimized camera poses ($T_0^*$ and $T_\tau^*$) within a pre-built 3D Gaussian map by jointly minimizing photometric, depth, and event residuals between rendered outputs and sensor observations when a new frame arrives. The rendering processes for both the image and the event are illustrated in~\Cref{fig:render}.

\textbf{Photometric Loss} We adopt the physical imaging process formulation addressing the limitations of conventional static exposure assumptions. The static model ignores how intensity changes over time during capture, which directly causes motion blur. Following~\cite{badgs, i2slam, xiong2024event3dgs}, digital camera imaging fundamentally involves two sequential stages: light capturing during sensor exposure with continuous photon accumulation over time, followed by photoelectric conversion that transforms the collected light into measurable electrical signals. This physical process is mathematically modeled as temporal integration of simulated latent sharp frames over the exposure duration and can be approximated by the discrete model as:
\begin{equation}\label{eq3}
    \begin{aligned}
       \bm{\tilde{I}}(\bm{u}) = \int_{0}^{\tau}  \bm{\mathcal{I}}(\mathbf{T}(\eta), \bm{u}) d\eta \approx \frac{1}{K} \sum_{k=0}^{K-1} \bm{\mathcal{I}}(\mathbf{T}(\eta_k), \bm{u})
    \end{aligned},
\end{equation}
where $\bm{\tilde{I}}(\bm{u})$ is integrated HDR image over the exposure time. $\tau$ is the exposure time, and $\bm{\mathcal{I}}(\mathbf{T}(\eta_k), \bm{u})$ specifies the instantaneous latent sharp image under pose $\mathbf{T}(\eta_k)$ derived by~\Cref{eq1} and~\Cref{eq2}. The timestamps $\eta_k$ are evenly divided based on the number of events, ensuring that each trajectory segment covers approximately the same distance.
Since event data inherently possesses HDR characteristics while images collected by the normal camera are 
represented in LDR, the scene representation is HDR. Inspired by~\cite{eventnerfslam}, we then introduce a Camera Response Function (CRF) to map the 
rendered HDR imagery into LDR space, thereby better preserving the joint characteristics of 
event data and image in tracking and mapping. Inspired by~\cite{jun2022hdr, i2slam},
we obtain the synthesized LDR image $\hat{\bm{I}}(\bm{u})$ by 
applying a trainable CRF to the rendered HDR image $\tilde{\bm{I}}(\bm{u})$:
\begin{equation}
    \small
    \begin{aligned}
         \hat{\bm{I}}(\bm{u}) &= {\rm CRF_{leaky}}(\tilde{\bm{I}}(\bm{u})) \\
         &=
        \begin{cases} 
        \alpha \tilde{\bm{I}}(\bm{u}),                              & \text{if }  \tilde{\bm{I}}(\bm{u}) < 0\\
        {\rm Interp}(\tilde{\bm{I}}(\bm{u}), \bm{Q})                    & \text{if } 0 \leq \tilde{\bm{I}}(\bm{u}) \leq 1,\\
        -\frac{\alpha}{\sqrt{\tilde{\bm{I}}(\bm{u})}} + \alpha + 1, & \text{if }   \tilde{\bm{I}}(\bm{u})>1
        \end{cases}
    \end{aligned}
\end{equation}
where ${\rm Interp}(\cdot)$ represents the linear interpolation function, $\bm{Q}$ means the trainable control nodes, and $\alpha$ is set as $0.01$ in our system. We uniformly fix $N$ output levels in the LDR space and associate each with a trainable HDR intensity, forming the control nodes $\bm{Q}$ and shared all frames. Given an HDR input $\tilde{\bm{I}}(\bm{u})$, its corresponding LDR output is obtained by linearly interpolating between adjacent control nodes, to approximate a differentiable CRF.
The photometric loss $L_I$ is then designed:
\begin{equation}
    \begin{aligned}
    L_{I} & =  \left\| \hat{\bm{I}}-\bm{I}_{obs} \right\|_1,
    \end{aligned}
\end{equation}
where $I_{obs}$ denote the LDR image acquired by the image camera, and $\left\| \cdot \right\|_1$ represents the $L_1$-norm operator.

\textbf{Event Loss} 
The event stream $\bm{\mathbb{E}} = \{e_m\}$ asynchronously captures spatiotemporal brightness changes, where each event $e_m = \{\bm{u}_m, t_m, p_m\}$ includes a pixel location $\bm{u}_m$, timestamp $t_m$, and polarity $p_m \in \{+1, -1\}$. An event is triggered when the logarithmic brightness change at pixel $\bm{u}_m$ exceeds a predefined threshold $\theta > 0$, i.e., $|\log(L(x, t_i + \Delta t)) - \log(L(x, t_i))| > \theta$. Due to their discrete and sparse nature, raw events are not directly suitable for training 3DGS. To address this, we aggregate events by position and polarity over short time intervals to construct a dense event map:
\begin{equation}
    \small
    \bm{E}_k(\bm{u}) = \sum_{e_m \in \mathcal{E}_k} p_m,
\end{equation}
where $\mathcal{E}_k = \{e_m \mid \bm{u}_m = \bm{u}, \eta_{k-1} < t_m < \eta_k\}$ is the subset of events within the $n$-th time window.
Since event maps cannot be directly rendered from the Gaussian scene representation, we simulate them using the event generation model~\cite{esim,xiong2024event3dgs} by computing the difference between the logarithmic brightness values of two consecutive rendered frames:
\begin{equation}
    \hat{\bm{E}}_k(\bm{u}) = \log(\hat{\bm{B}}(\bm{T}(\eta_k), \bm{u})) - \log(\hat{\bm{B}}(\bm{T}(\eta_{k-1}), \bm{u})),
\end{equation}
where $\hat{\bm{B}}$ denotes the grayscale brightness obtained from the rendered RGB image $\tilde{\bm{I}}$ via the BT.601 luma transform~\cite{601}.

The event loss is then defined as the L1-distance between the accumulated event map $\bm{E}_k$ and rendered event maps $\hat{\bm{E}}$:
\begin{equation}
    L_{HE} = \frac{1}{K} \sum_{n=0}^{K-1} \sum_{\bm{E}_k(\bm{u})\neq0} \left\| \theta \cdot \bm{E}_k(\bm{u}) - \hat{\bm{E}}_k(\bm{u}) \right\|_1.
\end{equation}

To further leverage event-based supervision, inspired by~\cite{enerf}, we define a no-event loss to penalize predicted undesired photometric changes at locations with no events:
\begin{equation}
    L_{NE} = \frac{1}{K}  \sum_{n=0}^{K-1} \sum_{\bm{E}_k(\bm{u})=0}  \left\| \hat{\bm{E}}_k(\bm{u}) \right\|_1.
\end{equation}
Unlike prior work~\cite{enerf}, we assume that in the absence of events, no photometric change has occurred. This assumption helps enforce temporal consistency and accelerates convergence in our GS-SLAM. The final event loss is designed as:
\begin{equation}
    L_{E} = L_{HE} + \lambda_{NE} L_{NE},
\end{equation}
where $\lambda_{NE}$ is the weighting factor for the no-event loss.

\renewcommand{\arraystretch}{1.1}
\begin{table*}
\centering
 \resizebox{0.92\linewidth}{!}{
\begin{tabular}{cccccccccc||c}
\toprule
\textbf{Method} & \textbf{Metric} & \textbf{room0} & \textbf{room1} & \textbf{room2} & \textbf{office0} & \textbf{office1} & \textbf{office3} & \textbf{office4} & \textbf{Avg.} & \textbf{Rendering FPS} \\
\hline
\multirow{3}{*}{PhotoSLAM~\cite{photoslam}} & PSNR[dB]$\uparrow$  & 18.61 & 19.66& \textbf{--} &  23.85& \textbf{--} & 17.74&  14.98& \textbf{--}   &  \multirow{3}{*}{1075.8}\\
                                        & SSIM$\uparrow$          & 0.569& 0.658& \textbf{--}& 0.710& \textbf{--}& 0.666& 0.616&\textbf{--}    &  \\
                                        & LPIPS$\downarrow$       & 0.390& \underline{0.385}& \textbf{--}& 0.347& \textbf{--}& 0.321& 0.456&\textbf{--}    &  \\ \hline
\multirow{3}{*}{MonoGS~\cite{monogs}} & PSNR[dB]$\uparrow$& 20.47 & 21.97 & 24.05 & 26.75& 26.76&21.41& 21.79& 23.32&  \multirow{3}{*}{\underline{1113.9}}\\
                                      & SSIM$\uparrow$& 0.632& 0.697& 0.768&  0.789& 0.830 & 0.749& 0.769& 0.748&  \\
                                      & LPIPS$\downarrow$ & 0.454& 0.451& 0.335& 0.365& 0.335& 0.267& 0.391& 0.371&   \\ \hline

\multirow{3}{*}{\makecell{MonoGS~\cite{monogs}\\(Refined)}} & PSNR[dB]$\uparrow$& \underline{21.20}& \underline{22.66} & \underline{24.43}& \underline{26.97}&                                                              \underline{27.12}& \underline{21.65}& \underline{23.31}&  \underline{23.91}&  \multirow{3}{*}{1052.7}\\
                                                & SSIM$\uparrow$&  \underline{0.651}& \underline{0.709}& \underline{0.777}& \underline{0.798}& \underline{0.836}& \underline{0.762}& \underline{0.798} & \underline{0.762}&  \\
                                                & LPIPS$\downarrow$ &  \underline{0.383}& 0.388& \underline{0.307} & \underline{0.320}& \underline{0.299}& \underline{0.249}& \underline{0.328}& \underline{0.325}&   \\ \hline

\multirow{3}{*}{\bf{Ours}} & PSNR[dB]$\uparrow$& \textbf{24.06}& \textbf{26.30}& \textbf{27.61}& \textbf{31.72}& \textbf{33.38}&\textbf{26.50}& \textbf{23.79}& \textbf{27.62}&  \multirow{3}{*}{\textbf{1134.2}}\\
& SSIM $\uparrow$& \textbf{0.744}& \textbf{0.783}& \textbf{0.838}&  \textbf{0.885}& \textbf{0.927}&\textbf{0.846}& \textbf{0.806}& \textbf{0.833}&  \\
& LPIPS$\downarrow$ & \textbf{0.229}& \textbf{0.256}&\textbf{0.172}&\textbf{0.142}& \textbf{0.123}&\textbf{0.113}& \textbf{0.242}& \textbf{0.182}&   \\

\bottomrule
\end{tabular}
}
\caption{\textbf{Rendering performance comparison of RGB-D SLAM methods on EventReplica.} It should be noted that, although Photoslam may experience tracking loss in some sequences, it has the ability to reinitialize itself. The results presented here incorporate the mapping outcomes subsequent to the reinitialization process. \textbf{--} means the reinitialization attempt did not succeed. Our method outperforms the existing methods.}
\label{table:ReplicaRender}
\vspace{-0.5em} 
\end{table*}

\begin{figure*}[t]
    \centering
    \includegraphics[width=0.8\linewidth]{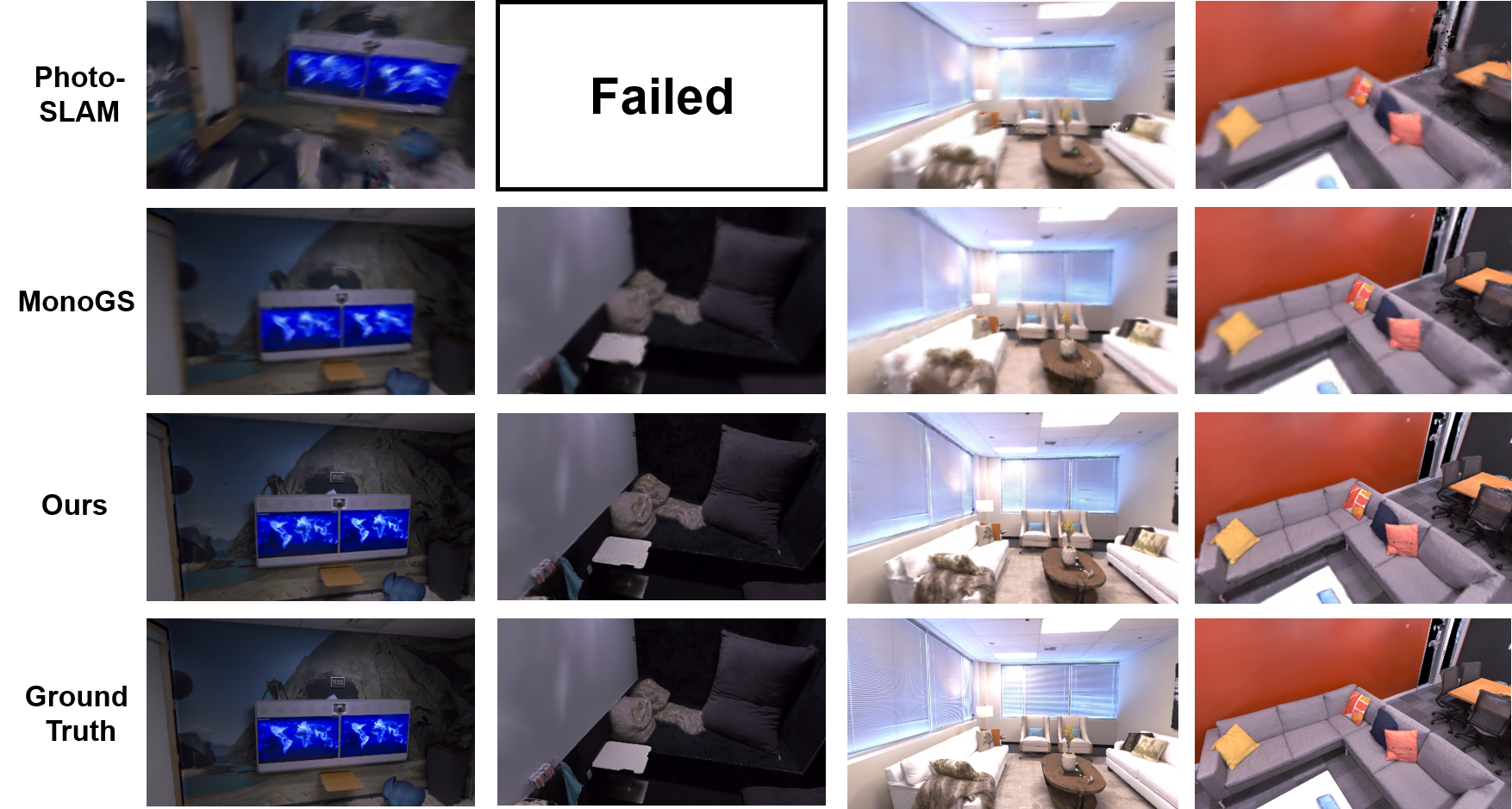}
    \caption{\textbf{Comparison of rendering quality on the EventReplica dataset.} Our method can achieve sharper reconstructions.}
    \label{fig:eventreplica_compare}
    \vspace{-0.8em} 
\end{figure*}

\textbf{Depth Loss}
The observed depth cannot be directly aligned with motion-blurred frames. Inspired by~\cite{i2slam}, we define the depth loss as the minimum discrepancy between the rendered depth $\mathcal{\bm{D}}$ and the sensor depth $\bm{D}_{\text{obs}}$ during the exposure: 
\begin{equation} L_D = \min_k \left\| \bm{D}_{\text{obs}} - \bm{\mathcal{D}}(\bm{T}(\eta_k)) \right\|_1. \end{equation} 
This formulation encourages alignment between the depth map and the latent sharp image within the exposure window.

\textbf{Pose Optimization}
To estimate the continuous trajectory during the current frame’s exposure, 
we optimize the control nodes while keeping the 3D Gaussian map $\mathcal{G}$ fixed. 
The optimized control nodes $T_0^*$ and $T_\tau^*$ are obtained by solving
the following objective through iterative optimization:
\begin{equation}
T_0^*, T_\tau^* = \arg \min_{T_0, T_\tau} \left( \lambda_E L_E + \lambda_{ID} \left( \lambda_I L_I + \lambda_D L_D \right) \right),
\end{equation}
where $ \lambda_E $, $ \lambda_I $, and $ \lambda_D $ are weights that control the individual contributions of each term, and $ \lambda_{ID} $ balances the combined image and depth terms relative to the event term.

\begin{table}[t]
\centering
\caption{\textbf{Comparison of tracking results ATE (cm) on EventReplica.} \textcolor{red}{\textbf{L}} means the method loses tracking in the sequence. * indicates odometry-only methods that do not support consistent or photorealistic 3D scene reconstruction.}
\label{table:eventreplica_ate}
\resizebox{0.95\linewidth}{!}{
\begin{tabular}{lcc||cc:c}
\toprule
\textbf{Scenes} & \makecell[c]{RampVO*\\~\cite{rampvo}} & \makecell[c]{ORB-SLAM2*\\~\cite{orbslam2}} & \makecell[c]{Photo-SLAM\\~\cite{photoslam}} & \makecell[c]{MonoGS\\~\cite{monogs}} & \textbf{Ours} \\
\midrule
\textbf{room0}    & \textbf{4.27} & 5.57 & \textbf{6.64} & 12.76 & \underline{6.70} \\
\textbf{room1}    & \textbf{13.11} & \textcolor{red}{\textbf{L}} & \textcolor{red}{\textbf{L}} & \underline{8.45} & \textbf{3.26} \\
\textbf{room2}    & \textbf{3.43} & \textcolor{red}{\textbf{L}} & \textcolor{red}{\textbf{L}} & \underline{3.64} & \textbf{3.16} \\
\textbf{office0}  & \textbf{4.41} & \textcolor{red}{\textbf{L}} & \textcolor{red}{\textbf{L}} & \underline{7.44} & \textbf{3.47} \\
\textbf{office1}  & \textbf{3.09} & \textcolor{red}{\textbf{L}} & \textcolor{red}{\textbf{L}} & \underline{7.78} & \textbf{3.53} \\
\textbf{office3}  & \textbf{4.33} & 6.48 & \underline{6.78} & 7.91 & \textbf{4.71} \\
\textbf{office4}  & \textbf{5.87} & \textcolor{red}{\textbf{L}} & \textcolor{red}{\textbf{L}} & \underline{16.55} & \textbf{12.41} \\
\hline
\textbf{Avg.}     & \textbf{5.50} & \textbf{-} & \textbf{-} & \underline{9.22} & \textbf{5.32} \\
\bottomrule
\end{tabular}
}
\end{table}

\subsection{Mapping Process}
The mapping process comprises two core components: keyframe management and 3DGS map updating. 

\textbf{Keyframe Management} After tracking converges, we follow~\cite{monogs} for keyframe selection and keyframe  window maintenance. A keyframe is inserted when the overlapped visible 3D Gaussians (IoU) with the latest keyframe falls below a threshold, promoting viewpoint diversity or camera movement exceeds a threshold scaled by the current frame's average depth. To maintain a bounded keyframe set $\mathcal{W}_k$ for mapping for computational efficiency, we remove historical keyframes whose overlap with the new keyframe is below another lower threshold to preserve local relevance. If no keyframes are removed, we adopt the strategy from~\cite{dso} to remove the most redundant keyframe to keep the window size fixed.

\textbf{3DGS Map Updating} After the creation of a novel keyframe, new Gaussians are instantiated and incorporated into the established 3DGS representation. Following~\cite{monogs}, the mean values $\bm{\mu}$ of the new Gaussians are initialized via depth back-projection leveraging estimated camera poses. The remaining parameters (including rotation $q$, scale $s$, opacity $o$, and color $c$) are initialized according to the strategy proposed in~\cite{gs}. We then need to optimize the parameters. Subsequent optimization of 3DGSs incorporates an isotropic regularization term~\cite{monogs} to mitigate artifacts caused by skinny Gaussians:
\begin{equation}
    \begin{aligned}
        L_{iso} = \frac{1}{|\mathcal{G}|} \sum_{i=1}^{|\mathcal{G}|} \left\| s_i - \bm{1} \cdot \overline{s}_i \right\|,
    \end{aligned}
\end{equation}
where $ \overline{s}_i$ denotes the mean of the scale $s_i$. Two randomly selected historical keyframes, $\mathbf{w}_r^1$ and $\mathbf{w}_r^2$, are added to the current window $\mathcal{W}_c$ to form the extended mapping set $\mathcal{W}' = \mathcal{W}_c \cup \{\mathbf{w}_r^1, \mathbf{w}_r^2\}$. The final mapping loss can be designed as:
\begin{equation}
    \begin{aligned}
        L_{map} = \lambda_{iso} L_{iso} + \sum_{w \in \mathcal{W}'} (\lambda_E L^{w}_E + \lambda_{ID} (\lambda_I L^{w}_I+ \lambda_D L^{w}_D))
    \end{aligned},
\end{equation}
where $L^{w}_I$, $L^{w}_E$, $L^{w}_D$ are the photometric loss, the event loss, and the depth loss of the frame $w$ and $\lambda_{iso}$ is taken as $10$. It is worth noting that the no-event loss plays a crucial role during mapping, as it significantly reduces the ringing problem—an artifact commonly observed in deblurring methods. 
We jointly optimize the parameters of the Gaussians and finetune the camera poses of the latest $k$ keyframes in a sliding window to improve consistency.
After the optimization of mapping, we prune the Gaussians for mapping stability.

\section{Experiments}
In this section, we first introduce the dataset we collected and used in our experimental setup. We then present qualitative visual comparisons with existing methods to demonstrate the superior performance of our approach in both tracking and mapping. Next, we provide additional visualizations of the rendered results, illustrating that our method outperforms existing approaches in reconstructing sharper and higher-quality 3DGS scenes. Finally, we conduct an ablation study on the contributions of event information, the Camera Response Function, the no-event loss, and the integration of GS-SLAM with single-frame deblurring, demonstrating the importance of each component in the overall effectiveness of our system.
\subsection{Dataset}

\textbf{EventReplica}
We created a synthetic event dataset, EventReplica, by extending the Replica dataset from~\cite{nice}. The original images were resized to 459×260 pixels and cropped to 448×256 pixels. Following~\cite{eventnerfslam, qi2023e2nerf, lu2025bardgsblurawarereconstructiondynamic}, we used FILM~\cite{reda2022film} to generate intermediate frames, which were then converted to event streams using VID2E~\cite{vid2e}. The physical motion of the frames is synthesized by summing all interpolated frames within the exposure time, with the final frame serving as the sharp ground truth and its depth map being directly adopted as the depth ground truth for the corresponding blurred frame.

\textbf{DEVD} 
Our data acquisition system consists of a DAVIS346 color event camera for capturing both events and images, a RealSense D435i depth camera for acquiring depth information, and the FZMotion Motion Capture System for providing ground-truth poses. Since both the D435i depth module and the motion capture system emit 850nm infrared light—which introduces significant noise to the event camera—we installed an infrared-cut filter (transmitting only the 400-700nm visible spectrum) in front of the event sensor. Our dataset comprises four scenes: Mahjong, Mountain, Table, and Testbed.

\begin{figure*}[t]
    \centering
    \includegraphics[width=0.8\linewidth]{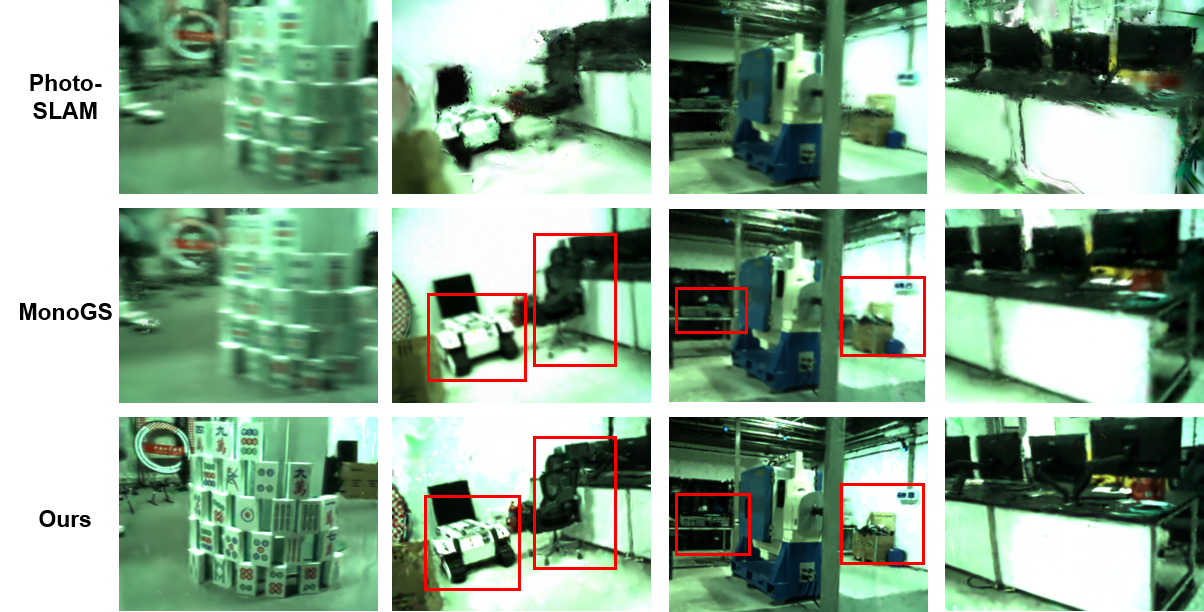}
    \caption{\textbf{Comparison of rendering quality on the DEVD dataset.} Our method can achieve sharper reconstructions.}
    \label{fig:devd_compare}
    \vspace{-0.8em}
\end{figure*}

\subsection{Implementation details}
We conducted our experiments on an NVIDIA RTX 4080 GPU. The event threshold $\theta$ was set to $0.25$ for synthetic data and $0.3$ for real data and $\lambda_{NE}=0.4$. We combined RGB and depth losses with weights $\lambda_I = 0.9$ and $\lambda_D = 0.1$. For the synthetic dataset, we set $\lambda_E = 0.05$ and $\lambda_{DI} = 0.95$; for the real dataset, $\lambda_E = 0.15$ and $\lambda_{DI} = 0.85$. A sliding window of size 10 was used for mapping, and the latest 5 frames were optimized in the backend. For fair comparison, we also set MonoGS to use the same window size. Other than that and the event-related parts, all settings are the same as MonoGS. \textbf{Boldface} indicates the best performing method and \underline{underline} indicates the second best in all experimental tables.

\begin{table}[t]
\centering
\caption{\textbf{Comparison of tracking results ATE (cm) on DEVD.} \ding{55} means the occurrence of some frame drops in this sequence. * indicates odometry-only methods that do not support consistent or photorealistic 3D scene reconstruction.}
\label{table:devd_ate}
\resizebox{0.95\linewidth}{!}{
\begin{tabular}{lcc||cc:c}
\toprule
\textbf{Scenes} & \makecell[c]{RampVO*\\~\cite{rampvo}} & \makecell[c]{ORB-SLAM2*\\~\cite{orbslam2}} & \makecell[c]{Photo-SLAM\\~\cite{photoslam}} & \makecell[c]{MonoGS\\~\cite{monogs}} & \textbf{Ours} \\
\midrule
\textbf{Mahjong1}   & \textbf{4.90} & 3.92\ding{55} & 4.34\ding{55} & \underline{3.17} & \textbf{1.45} \\
\textbf{Mahjong2}   & \textbf{3.77} & 3.72\ding{55} & 5.29 & \underline{2.55} & \textbf{1.19} \\
\textbf{Mountain1}  & \textbf{1.96} & 3.47 & 4.50 & \underline{4.30} & \textbf{1.18} \\
\textbf{Mountain2}  & \textbf{1.18} & 4.85 & 4.98 & \underline{3.52} & \textbf{1.70} \\
\textbf{Table1}     & \textbf{4.00} & 10.27 & 25.29 & \underline{6.73} & \textbf{2.97} \\
\textbf{Table2}     & \textbf{4.04} & 4.41 & \underline{4.21} & 10.77 & \textbf{6.56} \\
\textbf{Testbed1}   & \textbf{2.29} & 5.26 & \underline{3.69} & 5.15 & \textbf{2.66} \\
\textbf{Testbed2}   & \textbf{3.10} & 4.45 & \textbf{4.22} & 12.41 & \underline{7.58} \\
\hline
\textbf{Avg.}       & \textbf{3.16} & 5.04 & 7.06 & \underline{6.07} & \textbf{3.16} \\
\bottomrule
\end{tabular}
}
\end{table}

\subsection{Quantitative Evaluation}
In this section, we benchmark our method on EventReplica and DEVD, comparing it against existing GS-SLAM approaches, including Photo-SLAM~\cite{photoslam}, and MonoGS~\cite{monogs}. We further include a comparison with ORB-SLAM2~\cite{orbslam2}, a classical RGB-D SLAM system, and RampVO~\cite{rampvo}, the SOTA learning-based frame-event VO, to provide a comprehensive performance assessment. We use the single-thread mode of MonoGS~\cite{monogs} and evaluate Absolute Trajectory Error (ATE) over the entire trajectory, along with PSNR, SSIM, and LPIPS~\cite{gsslam, monogs} for reconstruction quality comparisons.

\textbf{Evaluation on synthesis dataset: EventReplica} Scene reconstruction quality comparisons are shown in~\Cref{table:ReplicaRender}, where the rendered images are evaluated against sharp ground-truth references. It is worth noting that although PhotoSLAM suffers from frequent and severe tracking loss, it can reinitialize and continue reconstruction; the results we report include such reinitialized reconstructions. MonoGS and PhotoSLAM still exhibit substantial performance degradation when reconstructing from motion-blurred images and are generally unable to recover clean and photorealistic scene appearances. Even when refined through offline reconstruction, MonoGS consistently fails to restore sharp and detailed content. In contrast, our method can significantly outperform the original MonoGS. Specifically, the PSNR improves from $23.32dB$ to $27.62 dB$ $(+4.20 dB)$, SSIM increases from $0.748$ to $0.833$ $(+0.085)$, and LPIPS drops from $0.371$ to $0.183$ $(–0.189)$. These improvements demonstrate the superior capability of our approach in producing sharp and clear 3DGS.

Trajectory comparisons are reported in~\Cref{table:eventreplica_ate}. Our method outperforms existing approaches in 6 out of 7 sequences, and We reduced the average error from $9.22cm$ to $5.32cm$, achieving an improvement of approximately $42.23\%$. ORB-SLAM2 and PhotoSLAM experience significant tracking degradation due to motion blur, which hampers keypoint detection and causes frequent mismatches. MonoGS also shows a notable drop in tracking performance, as it relies on the assumption of blur-free inputs and struggles under blurred conditions. RampVO, with its strong learning-based trackers, achieves comparable tracking performance to our method; however, unlike RampVO, which only produces a sparse point cloud lacking photometric information, our method enables photorealistic and high-quality clear scene reconstruction.

\textbf{Evaluation on real dataset: DEVD} Since ground-truth clean images are unavailable for the real-world dataset, direct quantitative comparisons are not feasible. Therefore, we present qualitative visual comparisons in the subsequent section to demonstrate the effectiveness of our method in recovering sharp and high-quality scene appearances. 

In~\Cref{table:devd_ate}, our method outperforms the existing baselines on the tracking performance, achieving the best ATE in 7 out of 8 sequences as well as the lowest average error overall. Compared to MonoGS, our approach yields a substantial improvement in tracking accuracy, with an average ATE reduction of approximately $47\%$ from $6.07cm$ to $3.16cm$. Although ORB-SLAM2 and Photo-SLAM incorporate re-localization mechanisms to recover from tracking failures, they still suffer from overall inferior performance compared to ours. RampVO achieves comparable tracking performance, but can only reconstruct a sparse point cloud, while our method enables photorealistic  and sharp scene reconstruction.

\begin{table*}[t]
    \centering
    \resizebox{0.95\linewidth}{!}{%
    \begin{tabular}{c c | c c c c : c c c c : c c c c}
        \toprule
        \multirow{2}{*}{\textbf{\makecell{Event \\ Tracking}}} &
        \multirow{2}{*}{\textbf{\makecell{Event \\ Mapping}}} &
        \multicolumn{4}{c}{\textbf{Room0}} &
        \multicolumn{4}{c}{\textbf{Room1}} &
        \multicolumn{4}{c}{\textbf{Office1}} \\
        \cline{3-14}
        & & ATE[cm]$\downarrow$ & PSNR[dB]$\uparrow$ & SSIM$\uparrow$ & LPIPS$\downarrow$ &
        ATE[cm]$\downarrow$ & PSNR[dB]$\uparrow$ & SSIM$\uparrow$ & LPIPS$\downarrow$ &
        ATE[cm]$\downarrow$ & PSNR[dB]$\uparrow$ & SSIM$\uparrow$ & LPIPS$\downarrow$\\ 
        % \hline
        \hline
        $\times$ & $\times$ & 11.61 & 19.99 & 0.618 & 0.321& 6.93 & 22.28& 0.704 & 0.348 & 7.07 & 28.13 & 0.837 & 0.198 \\ 
        % \hline
        $\times$ & $\checkmark$ & 11.47 & 19.88 & 0.617 & 0.310 & 6.84 & 21.81 & 0.688 & 0.346 & 8.02 & 27.35 & 0.827 & 0.199\\ 
        % \hline
        $\checkmark$ & $\times$ & \textbf{6.45} & \underline{23.14} &  \underline{0.714} & \underline{0.255} & \underline{4.75} & \underline{25.85} & \underline{0.768} & \underline{0.281} & \underline{3.60} & \underline{32.39} & \underline{0.914} & \underline{0.160} \\ 
        % \hline
       \textbf{$\checkmark$} & \textbf{$\checkmark$} & \underline{6.70} &\textbf{24.06}& \textbf{0.744} & \textbf{0.229} &\textbf{3.26} & \textbf{26.30} & \textbf{0.783} & \textbf{0.256} & \textbf{3.53} & \textbf{33.38} & \textbf{0.927} & \textbf{0.123} \\ 
        \bottomrule
    \end{tabular}
    }
    \caption{\textbf{The ablation study ATE(cm) analyzing the impact of event information on EventReplica.}}
    \label{tab:ablation_event_sync}
\end{table*}

\begin{table}[t]
    \centering
    \caption{\textbf{Ablation study on ATE (cm) evaluating the impact of event information on the DEVD.}}
    \resizebox{0.9\linewidth}{!}{
    \begin{tabular}{c c | c c c c}
        \toprule
        \scriptsize\textbf{\makecell{Event \\ Tracking}} &
        \scriptsize\textbf{\makecell{Event \\ Mapping}} &
        \textbf{Mahjong1} &
        \textbf{Mahjong2} &
        \textbf{Mountain1} &
        \textbf{Table1} \\
        \midrule
        $\times$ & $\times$ & 12.74 & 18.59 & 2.66 & 5.29 \\ 
        \hline
        $\times$ & $\checkmark$ & 14.77 & 6.54 & 2.56  & 5.18\\ 
        \hline
        $\checkmark$ & $\times$ & \underline{2.08} & \underline{1.63} & \underline{1.75}  & \underline{4.01}\\ 
        \hline
        \textbf{$\checkmark$} &\textbf{ $\checkmark$} & \textbf{1.45} & \textbf{1.19} & \textbf{1.18} & \textbf{2.97}\\ 
        \bottomrule
    \end{tabular}
    }
    \label{tab:ablation_event_real}
\end{table}

\begin{table}[t]
    \centering
    \caption{\textbf{The ablation study on ATE (cm) analyzing the effect of the CRF.}}
    \resizebox{0.85\linewidth}{!}{
    \begin{tabular}{c | c c c c}
        \toprule
         \textbf{Settings}&
        \textbf{Mahjong1} &
        \textbf{Mahjong2} &
        \textbf{Mountain1} &
        \textbf{Table1} \\
        \midrule
        w/o CRF     & 1.49  & 1.21 & 1.24 & 3.28  \\ 
        \textbf{w CRF (Ours)} & \textbf{1.45}  & \textbf{1.19} & \textbf{1.18} & \textbf{2.97}\\
        \bottomrule
    \end{tabular}
    }
    \label{tab:ablation_crf}
\end{table}

\textbf{Runtime analysis} The runtime analysis is carried out on the office1 sequence of the EventReplica dataset, where our method runs at 0.55 FPS and MonoGS reaches 1.75 FPS. The difference mainly comes from our design choice to perform more rendering times in order to combine events and images for improved performance. Importantly, this design enables our method to achieve robust tracking under challenging motion blur and to reconstruct sharp, high-fidelity 3D scenes—capabilities that MonoGS lacks despite its faster runtime. Overall, our approach prioritizes reconstruction quality and robustness, which are crucial for real-world deployment.

\vspace{-0.2em}
\subsection{Qualitative Evaluation}
In~\Cref{fig:eventreplica_compare} and~\Cref{fig:devd_compare}, we compare the mapping results of our method with PhotoSLAM and MonoGS. As shown, our rendered outputs can recover clean 3D images from blurry inputs, significantly outperforming the previous methods.

\vspace{-0.2em}
\subsection{Ablation Study}
\textbf{Event Information} We evaluate the contribution of event information to tracking and mapping in~\Cref{tab:ablation_event_sync} and~\Cref{tab:ablation_event_real}. We can see that when both event-based tracking and mapping are disabled, the system relies on multiple image renderings for direct tracking. This setup can be seen as a deblur-SLAM approach based solely on image and depth inputs~\cite{i2slam, mbaslam}. In such cases, the system struggles to recover sharp scene appearances from continuous motion blur and to achieve reliable 3D tracking. Although it may show marginal improvements over MonoGS, the overall performance remains limited.
Enabling event-based mapping without incorporating event tracking does not yield performance gains. This is primarily due to the absence of a robust tracking mechanism, which leads to inaccurate pose estimates. These inaccuracies degrade mapping quality, which in turn further hinders tracking performance—creating a negative feedback loop that significantly impairs the system’s overall effectiveness. In contrast, enabling event-based tracking without mapping results in a notable performance boost. This improvement stems from the precise pose estimation enabled by the event stream. With accurate poses, high-quality image reconstructions can be achieved using only the RGB frames and depth data. Our full model that employs both event-based tracking and mapping achieves the best overall performance, especially on real-world datasets.

\begin{figure}[t]
    \centering
    \includegraphics[width=\linewidth]{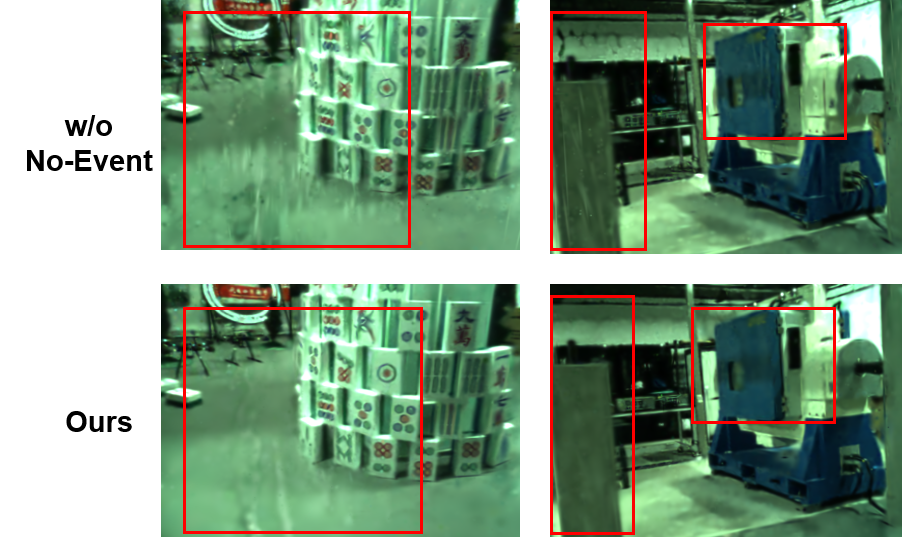}
    \caption{\textbf{The ablation study on the No-Event Loss.} The no-event loss is highly effective in removing ringing artifacts that resemble ripple patterns.}
    \label{fig:ringing}
\end{figure}

\begin{table}[t]
    \centering
    \caption{\textbf{An ablation study on single-image deblurring.}}
    \resizebox{0.95\linewidth}{!}{
    \begin{tabular}{c| c c c | c c c}
        \toprule
        \textbf{Method} & \multicolumn{3}{c|}{\textbf{Room0}} & \multicolumn{3}{c}{\textbf{Room1}}\\
        \cline{2-7} 
        & ATE[cm]$\downarrow$ & PSNR[dB]$\uparrow$ & LPIPS$\downarrow$ & ATE[cm]$\downarrow$ & PSNR[dB]$\uparrow$ & LPIPS$\downarrow$\\ 
        \hline
        MonoGS & 12.76 & 20.47 & 0.454 &  8.45& 21.97 & 0.451\\
        MonoGS+EDI & 16.96 & 18.18 & 0.429 & 11.6 & 19.67 & 0.446 \\
        \hline
        \textbf{Ours} & \textbf{6.70} & \textbf{24.06} & \textbf{0.229} &\textbf{3.26} & \textbf{26.30} & \textbf{0.256}  \\
        \bottomrule
    \end{tabular}
    }
    \label{tab:ablation_edi}
\end{table}

\textbf{No Event Loss} 
We observe that in real-world scenarios where depth quality is limited, ringing artifacts—characterized by ripple-like distortions—easily appear in the rendered images. In~\Cref{fig:ringing}, we compare the effectiveness of the no-event loss and find that it substantially suppress these artifacts.

\textbf{Camera Response Function} We evaluate the impact of the Camera Response Function (CRF) on the Majhong1, Mahjong2, Mountain1, and Table1 sequences. As shown in~\Cref{tab:ablation_crf}, the performance degrades when the CRF is not applied. This degradation is primarily due to the inherent difference in dynamic ranges between the event data and frame images in real-world datasets. Without CRF calibration, forcing the two modalities into a shared range leads to substantial inconsistencies. By introducing the CRF, we achieve better alignment between modalities, reducing these inconsistencies and enhancing overall system performance. 

\textbf{Single-Frame Deblur} 
For single-frame deblurring, we incorporate a widely used method, EDI~\cite{edi}, into MonoGS~\cite{monogs}. As shown in~\Cref{tab:ablation_edi}, the results reveal that the performance degrades after applying EDI, even compared to the baseline without deblurring. This degradation arises from artifacts introduced by single-frame deblurring methods, which act as noise and adversely affect the tracking and mapping results.

\section{Conclusion}
In this paper, we presented the first E-RGB-D Gaussian Splatting SLAM framework that integrates event data with image and depth inputs to enable robust tracking and high-fidelity mapping under motion blur. By explicitly modeling the camera's continuous trajectory during exposure and introducing a learnable CRF, our method effectively bridges the temporal and dynamic range discrepancies between asynchronous event streams and conventional image frames. Additionally, we proposed a no-event loss to suppress ringing artifacts, further improving the reconstruction quality. Extensive evaluations on both synthetic and real-world datasets demonstrate that our approach consistently outperforms existing GS-SLAM baselines in terms of localization accuracy and 3D scene fidelity. As the current system relies on depth input for both tracking and mapping, future work will focus on extending the framework to monocular setups.

\bibliography{reference.bib}{}
\bibliographystyle{IEEEtran}

\end{document}